\documentclass[11pt]{article}

\usepackage[final]{acl}

\usepackage{times}
\usepackage{latexsym}

\usepackage[T1]{fontenc}

\usepackage[utf8]{inputenc}

\usepackage{microtype}

\usepackage{inconsolata}

\usepackage{graphicx}
\usepackage{booktabs}
\usepackage[table]{xcolor}
\usepackage{siunitx}
\usepackage{threeparttable}
\usepackage{makecell}
\usepackage{amssymb}
\usepackage{adjustbox}
\usepackage{amsmath}
\usepackage{amssymb}
\usepackage{booktabs,array,listings}
\definecolor{groupgray}{HTML}{F4F5F7}
\definecolor{oursblue}{HTML}{EAF2FB}
\definecolor{expertgray}{HTML}{F4F1EC}

\newcommand{\yes}{\ensuremath{\checkmark}}
\newcommand{\no}{\ensuremath{\times}}
\newcommand{\resultfont}{\tiny}

\newcommand{\best}[1]{\multicolumn{1}{c}{{\resultfont\textbf{#1}}}}
\newcommand{\second}[1]{\multicolumn{1}{c}{{\resultfont\underline{#1}}}}

\title{From Collaboration to Capability: Internalizing Routed LLM Experts into Compact Reasoners}

\author{
  \textbf{Frank Nie}\textsuperscript{1,*},
  \textbf{Shuyao Wang}\textsuperscript{2,*},
  \textbf{Ethan B. Liu}\textsuperscript{1,*,\textdagger}
\\
\\
\textsuperscript{1}Shandong University
\\
  \textsuperscript{2}Zhejiang University
\\
{\footnotesize
  \textsuperscript{*}Equal contribution.
  \quad
  \textsuperscript{\textdagger}Corresponding authors.
}
}

\begin{document}
\maketitle
\begin{abstract}
A compact controller can coordinate stronger experts by selecting whom to consult, formulating requests, and integrating their responses. We study whether learning from both the controller's decisions and the experts' reasoning and code improves its generation after expert removal. We introduce \textsc{Rivet} for \emph{collaboration internalization}: expert-augmented reinforcement learning applies a shared outcome signal to controller decisions and returned expert spans, and verified trajectory internalization consolidates complete successful interactions through format-aware supervised training. The deployed controller generates reasoning, code, and interaction structure with local Python execution and no external LLM. Across seven competition-mathematics benchmarks, RIVET-1.7B and RIVET-4B achieve average accuracies of $28.25\%$ and $44.16\%$; Stage~II improves RIVET-4B's accuracy after expert removal by $6.49$ points, and GPQA-Diamond results provide evidence of generalization to scientific reasoning. Ablations show gains from ordinary trajectory supervision and additional format weighting, supporting the effectiveness of training on the content and structure of verified collaborations.
\end{abstract}

\section{Introduction}
\label{sec:introduction}

\begin{figure}[t]
\centering
\includegraphics[width=\columnwidth]{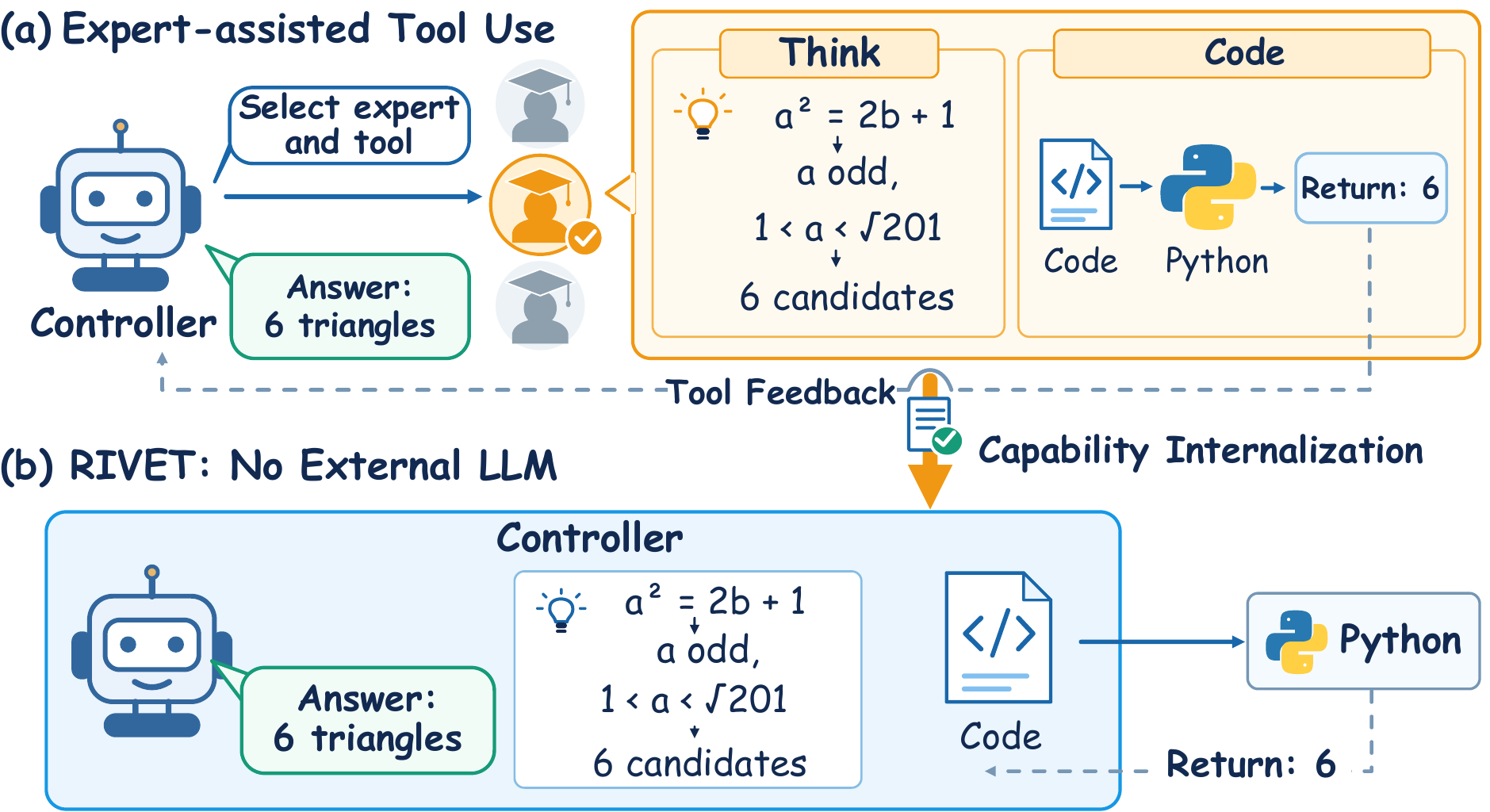}
\caption{From coordinating experts to generating their contributions.
(a) The controller requests expert reasoning and code; Python verifies six valid triangles.
(b) Schematic of internalized inference: the controller generates reasoning and code with local Python and no external LLM.
Training uses both controller decisions and expert content; Appendix~\ref{app:stage2-conversion-example} gives the source trajectory and its conversion into supervision.}
\label{fig:intro}
\end{figure}

Compact language models can extend their reasoning through executable tools and stronger language-model experts~\citep{feng2025retoolreinforcementlearningstrategic,li2025torlscalingtoolintegratedrl,cheng2026teachingthinkingmodelsreason,yu2025demystifyingreinforcementlearningagentic}. A controller coordinates this assistance by choosing experts, formulating requests, and integrating the returned content. Figure~\ref{fig:intro}(a) illustrates a controller counting integer right triangles with legs $a,b$, hypotenuse $b+1$, and $b<100$: it requests a derivation and then code to verify the count of six. The interaction records both the controller's decisions and the expert's contributions. \emph{What if a controller could learn not only how to coordinate experts, but also how to perform the reasoning it delegates to them?} Figure~\ref{fig:intro}(b) illustrates this goal, with the controller generating reasoning and code while local Python executes the code.

Existing approaches learn reasoning content and problem-solving behavior through several forms of supervision. Model routing and multi-agent orchestration select and combine LLMs for inference-time quality or cost~\citep{ong2024routellm,wang2024mixtureofagents,zhang2025routerr1,su2025toolorchestra}. Agent Distillation transfers reasoning and tool-use behavior through teacher-generated trajectories~\citep{kang2025distillingllmagentsmall}. On-policy distillation provides teacher guidance at student-visited states~\citep{agarwal2024onpolicydistillationlanguagemodels,tan2026atodannealedturnawareonpolicy}, with SOD adapting its strength across tool-use steps~\citep{zhong2026sodstepwiseonpolicydistillation}. Our focus is the interaction between a controller and the heterogeneous experts it requests: how to train on both the decisions that elicit assistance and the returned content within the same collaborative history.

A collaborative trajectory contains two connected sources of supervision. Its \emph{control decisions} specify the expert, request, and interaction point; its \emph{expert content} supplies the reasoning and code returned in that context. Our key idea is to train on both sides of the interaction, preserving the sequence that connects requested assistance to subsequent decisions. We call this \emph{collaboration internalization}, and evaluate whether it improves generation with external experts removed and local Python retained. This evaluation complements collaborative accuracy: our Stage-I 4B controller reaches $46.22\%$ average mathematics accuracy with experts and $37.67\%$ without them (Tables~\ref{tab:main-results} and~\ref{tab:expert-ablation}).

Learning from control decisions and expert content presents two challenges. First, \emph{updating both sides of collaboration} requires assigning a learning signal to externally generated spans as well as controller decisions. Updating only controller tokens leaves returned expert content outside the direct training objective. Second, \emph{consolidating their sequential relationship} requires training on requests, responses, and subsequent decisions as a complete interaction. Supervision must cover both reasoning content and the structure that supports code execution, while distinguishing content the controller should generate from observations that the environment must supply.

We introduce \textsc{Rivet} (\textbf{R}outing and \textbf{I}nternalization via \textbf{V}erified \textbf{E}xpert \textbf{T}rajectories) to train on the control decisions and expert content of collaborative interactions. Stage~I, \emph{expert-augmented GRPO}, applies a shared outcome advantage to controller decisions and the expert reasoning and code they elicit (Section~\ref{sec:stage1}). Stage~II, \emph{verified trajectory internalization}, further consolidates complete, correct, protocol-valid collaborations through supervised training with additional weight on tool-call structure (Section~\ref{sec:stage2}). In the triangle example, prediction targets include the requests and subsequent controller continuation as well as the expert's derivation and verification code; actual execution results remain environment observations. Appendix~\ref{app:stage2-conversion-example} illustrates this conversion. At deployment, one model generates the interaction sequence, with pauses for local Python execution.

We evaluate RIVET at two controller scales on seven competition-mathematics benchmarks and test generalization to scientific reasoning on GPQA-Diamond. Our contributions are:
\begin{itemize}
    \item We introduce RIVET, a framework that trains a controller on both the decisions that organize expert assistance and the content experts provide, targeting generation after expert removal (Section~\ref{sec:method}).
    \item We connect outcome-weighted updates on controller and expert spans with format-aware consolidation of complete verified interactions, preserving requests, expert responses, and subsequent decisions as prediction targets while retaining execution feedback as context (Sections~\ref{sec:stage1} and~\ref{sec:stage2}).
    \item We quantify the gap between assisted and independent generation and show a $6.49$-point Stage-II gain in RIVET-4B's mathematics accuracy after expert removal. Ablations measure the gains from trajectory supervision and format weighting, and compare the complete pipeline under different collaboration policies (Section~\ref{sec:experiments}).
\end{itemize}

\section{Related Work}
\label{sec:related_work}

\paragraph{Tool-integrated reasoning and agentic reinforcement learning.}
Tool-integrated reasoning augments language-model reasoning with executable tools and iterative feedback. ReTool uses cold-start SFT followed by outcome-based RL for code-interpreter use~\citep{feng2025retoolreinforcementlearningstrategic}; ToRL learns multi-turn code use directly from final-answer rewards~\citep{li2025torlscalingtoolintegratedrl}; and TRICE combines tool-integrated SFT and RL~\citep{cheng2026teachingthinkingmodelsreason}. DemyAgent emphasizes end-to-end trajectories and exploration-oriented optimization~\citep{yu2025demystifyingreinforcementlearningagentic}, while ARPO and AEPO modify sampling, advantage attribution, or entropy control~\citep{dong2025agenticreinforcedpolicyoptimization,dong2025agenticentropybalancedpolicyoptimization}. Recent studies further analyze collapse and unnecessary tool use in long-horizon agents~\citep{hao2026multisteptoolusereinforcementlearning,chen2026learningactmitigatingtool}. This line primarily studies a policy interacting with functional tools; RIVET additionally treats other language models as selectable sources of task-solving supervision.

\paragraph{Model routing and training-time guidance.}
Model routers select among stronger and weaker LLMs~\citep{ong2024routellm}, and multi-agent systems aggregate responses from several models~\citep{wang2024mixtureofagents}. Router-R1 learns multi-round model invocation and aggregation through RL while masking routed-model outputs from its training loss~\citep{zhang2025routerr1}; ToolOrchestra trains a lightweight model to coordinate heterogeneous models and tools using outcome and efficiency rewards~\citep{su2025toolorchestra}. Several methods provide denser training-time guidance: PACT uses fixed privileged traces~\citep{du2026pactprivilegedtracecotraining}, OPID extracts hindsight skills from completed policy trajectories~\citep{yang2026opidonpolicyskilldistillation}, and ATOD combines teacher distributions with reward optimization~\citep{tan2026atodannealedturnawareonpolicy}. RIVET includes the expert responses requested during collaborative solving in the controller's update, then consolidates verified interactions for deployment without those experts.

\paragraph{Trajectory construction and distillation.}
Tool-Star synthesizes multi-tool trajectories through tool-integrated prompting and filtering~\citep{dong2025toolstarempoweringllmbrainedmultitool}; AgentMath converts natural-language solutions into executable tool-augmented trajectories~\citep{luo2026agentmathempoweringmathematicalreasoning}; and COVERT constructs tool-use environments with multi-level verification~\citep{xu2026controllableverifiabletoolusedata}. Tool-R0 co-evolves task generation and tool-using policies through self-play~\citep{acikgoz2026toolr0selfevolvingllmagents}. Agent Distillation transfers reasoning, retrieval, and code-use trajectories from a large teacher agent~\citep{kang2025distillingllmagentsmall}, while Knowledge Purification selects or consolidates rationales from multiple teachers~\citep{jin2026knowledgepurification}. QR-Distill routes filtered reasoning paths to students according to their learning states~\citep{lei2025qrdistill}. On-policy distillation trains on student-generated sequences using teacher feedback~\citep{agarwal2024onpolicydistillationlanguagemodels}; SOD adapts that guidance across tool-use steps according to student--teacher divergence~\citep{zhong2026sodstepwiseonpolicydistillation}. RIVET uses responses elicited by a controller's requests to heterogeneous experts in outcome-weighted updates and subsequent verified-trajectory training.

AgentArk distills reasoning processes from homogeneous multi-agent debates~\citep{luo2026agentarkdistillingmultiagentintelligence}, whereas RIVET trains on controller-requested heterogeneous expert interactions with executable tools, including both requesting decisions and returned expert content.

\begin{figure*}[t]
    \centering
    \includegraphics[width=\textwidth]{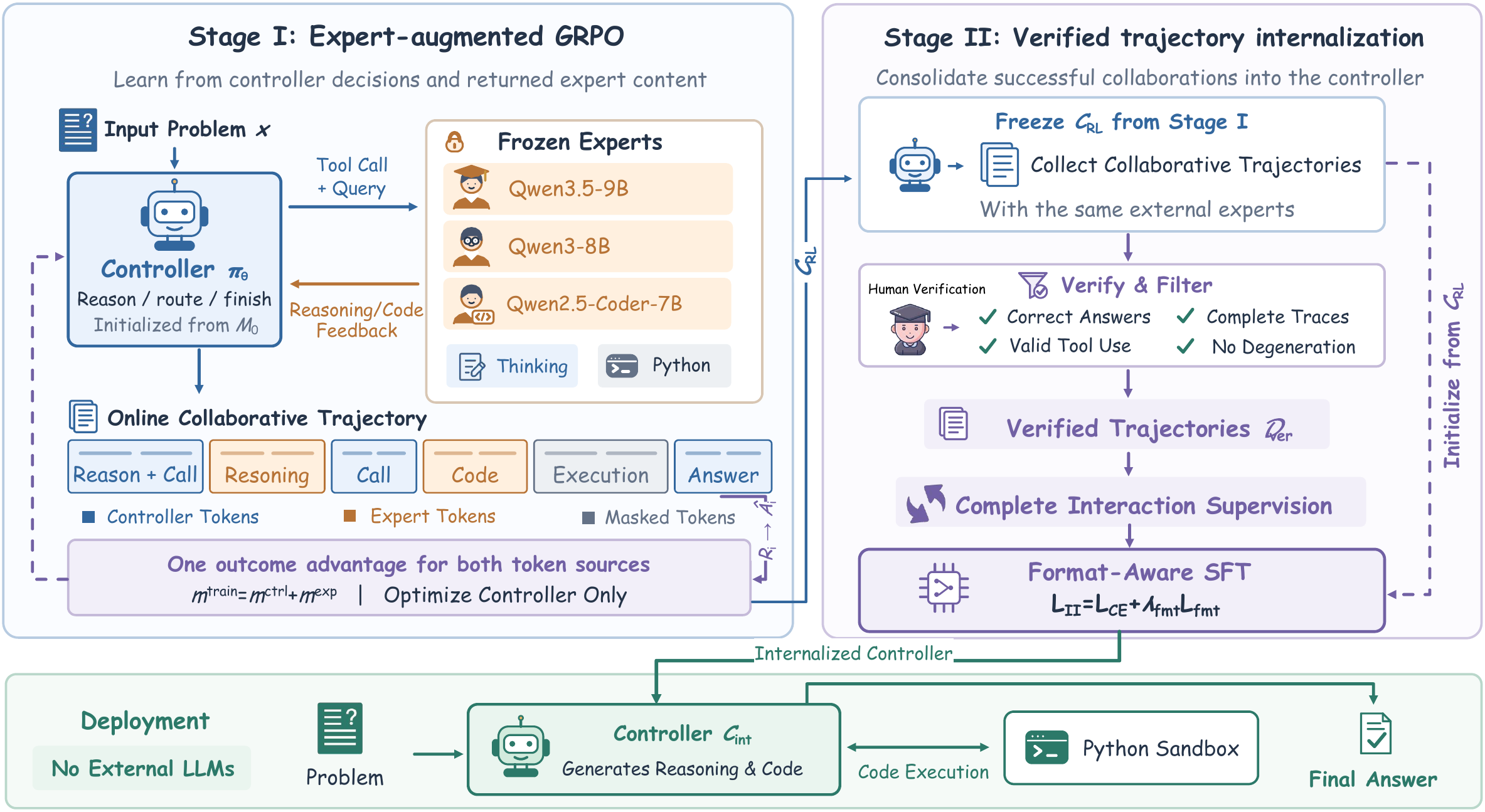}
    \caption{Training overview of RIVET. Stage~I trains the controller on its decisions and returned expert reasoning and code using a shared outcome signal. Stage~II consolidates verified successful interactions through format-aware supervised fine-tuning. The deployed controller generates its own reasoning and code, with local Python execution and no external LLMs.}
    \label{fig:rivet-overview}
\end{figure*}

\section{Method}
\label{sec:method}

\subsection{Overview}
\label{sec:method_overview}

\textsc{Rivet} trains on two connected parts of a collaborative trajectory: controller decisions that organize expert assistance and expert content returned in response (Figure~\ref{fig:rivet-overview}). Control decisions include expert selection, requests, and continuation or termination; expert content comprises the returned reasoning and code. Stage~I uses the trajectory outcome to update the controller on both parts. Stage~II freezes the learned collaboration policy to collect verified successful interactions, then consolidates their control decisions and expert content in context through format-aware supervised training on the complete sequences. At deployment, a single controller generates the reasoning, code, and interaction structure; actual Python execution results remain environment observations.

\subsection{Problem Formulation}
\label{sec:problem_formulation}
Given a problem $x$ with reference answer $y^\star$, let $\pi_\theta$ denote a controller initialized from a base model $M_0$. During collaboration, the controller can reason independently or invoke a tool from a unified set $\mathcal{U}$ backed by frozen experts $\mathcal{E}=\{E_k\}_{k=1}^{K}$. Each expert exposes two tools: a thinking tool that returns reasoning, and a Python tool that requests expert-generated code, executes it in a sandbox, and returns execution feedback. At interaction step $s$, the controller selects its next action from the history $h_s$:
\begin{equation}
\small
\begin{aligned}
a_s &\sim \pi_\theta(\cdot\mid h_s),a_s &\in \{\mathrm{reason},\mathrm{call}(u,q_s),\mathrm{finish}\},
\end{aligned}
\label{eq:action_space}
\end{equation}
where $u\in\mathcal{U}$ identifies the selected tool and $q_s$ is the instruction generated by the controller. Tool outputs are appended to the history before the controller continues. The trajectory ends when the controller returns a final answer. The controller's history $h_s$ conditions its choice of tool and request; the selected pair $(u,q_s)$ determines the expert source and requested content of the resulting supervision.

\subsection{Stage I: Expert Augmented GRPO}
\label{sec:stage1}

Stage~I learns from expert content requested during the controller's own solving process. Selecting an expert, forming a query, and choosing when to continue or stop determine both the available assistance and the responses encountered in training. Expert-augmented GRPO includes those responses in the controller's update by assigning controller and expert spans the same trajectory-outcome advantage. The reward optimizes solution correctness; it does not directly measure a response's value for later internalization.

\paragraph{Outcome-based reward.} Following \citet{li2025torlscalingtoolintegratedrl}, we reward final-answer correctness and assign no intrinsic bonus to tool calls. Calls therefore receive credit through the outcome of the complete trajectory. For each problem, we sample $G$ trajectories $\{\tau_i\}_{i=1}^{G}$ and assign the following reward to the final answer $\hat y_i$ of trajectory $\tau_i$:

\begin{equation}
\small
R_i=
\begin{cases}
+1, & \operatorname{Verify}(\hat y_i,y^\star)=1,\\
-1, & \text{otherwise}.
\end{cases}
\label{eq:outcome_reward}
\end{equation}

We use Qwen's native tool-calling schema and rely on the base model's pretrained format-following capability, without a separate format reward.

\paragraph{Expert-augmented GRPO.} Sparse final-answer rewards provide limited guidance over long collaborative trajectories, making exploration vulnerable to premature policy entropy collapse~\citep{yu2025demystifyingreinforcementlearningagentic,dong2025agenticentropybalancedpolicyoptimization}. RIVET therefore includes returned expert reasoning and code in the update. Let $y_{i,t}$ be token $t$ in the serialized trajectory $\tau_i$, with prefix $h_{i,t}=(x_i,y_{i,<t})$. The disjoint masks $m_{i,t}^{\mathrm{ctrl}}$ and $m_{i,t}^{\mathrm{exp}}$ select controller-generated tokens and external-expert reasoning or code, respectively. Actual Python execution outputs are environment observations and are assigned zero in both masks. We form the group-relative advantage

\begin{equation}
\small
A_i=\frac{R_i-\mu_G}{\sigma_G+\epsilon_A},
\label{eq:group_advantage}
\end{equation}

where $\mu_G$ and $\sigma_G$ are the mean and standard deviation of the $G$ rewards for the same problem. The scalar $A_i$ is broadcast to every trainable token in $\tau_i$. Let $\theta_{\mathrm{old}}$ denote the controller before the current update. After the complete collaborative trajectory has been collected, we recompute its token probabilities by teacher forcing under this frozen pre-update controller and define

\begin{equation}
\small
\rho_{i,t}(\theta)=
\frac{\pi_\theta(y_{i,t}\mid h_{i,t})}
{\pi_{\theta_{\mathrm{old}}}(y_{i,t}\mid h_{i,t})}.
\label{eq:expert_token_ratio}
\end{equation}

For $s(\rho,A)=\min\!\left(\rho A,\operatorname{clip}(\rho,1-\epsilon,1+\epsilon)A\right)$, the dual-clipped surrogate is

\begin{equation}
\small
g(\rho,A)=
\begin{cases}
s(\rho,A), & A\geq 0,\\
\max\!\left(s(\rho,A),cA\right), & A<0.
\end{cases}
\label{eq:dual_clip_surrogate}
\end{equation}

With $m_{i,t}^{\mathrm{train}}=m_{i,t}^{\mathrm{ctrl}}+m_{i,t}^{\mathrm{exp}}$ and $N_{\mathrm{train}}=\sum_{i,t}m_{i,t}^{\mathrm{train}}$, Stage~I minimizes

\begin{equation}
\small
\mathcal{L}_{\mathrm{I}}(\theta)
=-\frac{1}{N_{\mathrm{train}}}
\sum_{i,t}m_{i,t}^{\mathrm{train}}
g\!\left(\rho_{i,t}(\theta),A_i\right).
\label{eq:stage1_loss}
\end{equation}

We use token-mean normalization without an additional KL term. For an expert-origin token, the denominator in Eq.~\eqref{eq:expert_token_ratio} is its teacher-forced likelihood under the pre-update \emph{controller}, rather than under the frozen expert that produced it. The resulting term performs advantage-weighted online distillation with the same proximal ratio used for controller tokens; it is not an importance-sampling estimate for the expert policy. Controller-token updates learn reasoning, routing, refinement, and termination, while expert-token updates transfer reasoning and code observed at controller-visited states. We denote the resulting controller by $C_{\mathrm{RL}}=\pi_{\theta_{\mathrm{RL}}}$.

\subsection{Stage II: Verified Trajectory Internalization}
\label{sec:stage2}

Stage~II consolidates complete successful collaborations to train reasoning, code, and tool-call generation after expert removal. Stage~I already updates the controller on expert spans, while receiving those spans from external models during rollouts and learning from both successful and unsuccessful trajectories. Stage~II freezes the learned policy to collect successful interactions and supervises their verified sequences, retaining Python execution outputs as environment context.

\paragraph{Verified trajectory construction.} We freeze $C_{\mathrm{RL}}$ and generate eight collaborative trajectories per training problem using the same tool set. We retain only complete, protocol-compliant trajectories with correct, parseable final answers, valid Python execution when applicable, and no severe repetition, malformed text, or degenerate call loops. Both tool-assisted and independently solved trajectories are eligible. Following prior work that penalizes excessive tool calls~\citep{chen2026learningactmitigatingtool}, we prioritize candidates with fewer calls when multiple trajectories satisfy all quality criteria. After deduplication and manual review, we retain at most two trajectories per problem. The resulting internalization datasets $\mathcal{D}_{\mathrm{ver}}$, generated entirely from the training split, contain 12,000 trajectories for Qwen3-4B-Instruct and 10,000 for Qwen3-1.7B.

\paragraph{Training the controller to generate complete interactions.} We preserve Qwen's native tool-call serialization and canonicalize missing or free-form instructions using fixed tool-specific templates. The supervised sequence includes the controller's requests, the expert reasoning and code they elicited, and the final answer as prediction targets. For the resulting sequence $z_{1:T}$, $m_t^{\mathrm{sft}}=1$ for controller-generated spans, tool-call structure, expert reasoning or code, and the final answer, whereas problem tokens and Python execution outputs have $m_t^{\mathrm{sft}}=0$. In the triangle example, this places both the derivation and the verification program in the controller's target sequence. At internalized inference, the controller generates these kinds of content in the same decoding stream, pausing for local Python execution when it generates code. Appendix~\ref{app:stage2-conversion-example} provides the exact templates, token ownership, execution interface, and complete before--after training example.

\paragraph{Format-aware supervised internalization.} Starting from $C_{\mathrm{RL}}$, we fine-tune the controller on the canonicalized $\mathcal{D}_{\mathrm{ver}}$. Let $N_{\mathrm{sft}}=\sum_{t=1}^{T}m_t^{\mathrm{sft}}$. The masked next-token cross-entropy loss is

\begin{equation}
\small
\mathcal{L}_{\mathrm{CE}}
=-\frac{1}{N_{\mathrm{sft}}}\sum_{t=1}^{T}
m_t^{\mathrm{sft}}\log\pi_\theta(z_t\mid x,z_{<t}).
\label{eq:stage2_ce}
\end{equation}

We place additional weight on the format tokens that specify each native tool call. Let $\Omega_{\mathrm{fmt}}\subseteq\{t:m_t^{\mathrm{sft}}=1\}$ denote their positions, including tool-call delimiters, tool identifiers, and required argument fields. The format loss is

\begin{equation}
\small
\mathcal{L}_{\mathrm{fmt}}
=-\frac{1}{|\Omega_{\mathrm{fmt}}|}
\sum_{t\in\Omega_{\mathrm{fmt}}}
\log\pi_\theta(z_t\mid x,z_{<t}).
\label{eq:stage2_format}
\end{equation}

For trajectories without tool calls, we set $\mathcal{L}_{\mathrm{fmt}}=0$. This term supervises the structure of calls present in verified trajectories; it supplies no separate reward for choosing to invoke a tool or for call frequency. The combined loss is

\begin{equation}
\small
\mathcal{L}_{\mathrm{II}}
=\mathcal{L}_{\mathrm{CE}}
+\lambda_{\mathrm{fmt}}\mathcal{L}_{\mathrm{fmt}},
\label{eq:stage2_sft}
\end{equation}

where $\lambda_{\mathrm{fmt}}$ controls the additional weight on tool-call structure. Training with Eq.~\eqref{eq:stage2_sft} yields the internalized controller $C_{\mathrm{int}}=\pi_{\theta_{\mathrm{int}}}$. Its supervision includes the reasoning, code, and interaction structure of verified collaborations as well as their final answers.

\paragraph{Inference modes.} We evaluate \textsc{Rivet} in two settings. In \emph{collaborative mode}, the Stage-I controller $C_{\mathrm{RL}}$ uses the expert-backed tools and provides an expert-assisted reference. In \emph{internalized mode}, the Stage-II controller $C_{\mathrm{int}}$ invokes no external LLM; a local Python sandbox executes code generated by the controller. Internalized mode is the primary deployment setting. For the expert-removal diagnostic, we also evaluate $C_{\mathrm{RL}}$ with external experts disabled and local Python available. Comparing this model with $C_{\mathrm{int}}$ measures the change after Stage~II under the same deployment interface.

\section{Experiments}
\label{sec:experiments}

Our evaluation tests the effectiveness of training on controller decisions and expert content for generation after expert removal. Main results establish performance on competition mathematics and scientific reasoning. Ablations then address three questions: how much accuracy is lost when experts are removed, which Stage-II training choices improve performance under that interface, and how the collaboration policy affects the final two-stage model. Retention/recovery diagnostics characterize changes in solved instances, while routing and call-budget analyses describe control behavior during expert-assisted inference.

\begin{table*}[t]
\centering
\caption{Main results on math benchmarks. Bold and underlined values denote the best and second-best results. Expert-assisted variants are references excluded from the ranking. SOD-Teacher-4B denotes the GRPO-trained teacher released with SOD.}
\label{tab:main-results}
\begin{threeparttable}
\renewcommand{\resultfont}{\scriptsize}
\small
\setlength{\tabcolsep}{4pt}
\renewcommand{\arraystretch}{1.10}
\scalebox{1}{%
\begin{tabular}{lc *{8}{>{\resultfont}S[table-format=2.2]}}
\toprule
Method
& \makecell{Tool}
& \multicolumn{1}{c}{\makecell{AIME\\2024}}
& \multicolumn{1}{c}{\makecell{AIME\\2025}}
& \multicolumn{1}{c}{\makecell{AIME\\2026}}
& \multicolumn{1}{c}{\makecell{HMMT\\2025}}
& \multicolumn{1}{c}{\makecell{Beyond\\AIME}}
& \multicolumn{1}{c}{\makecell{IMO\\Answer}}
& \multicolumn{1}{c}{\makecell{APEX\\2025}}
& \multicolumn{1}{c}{\makecell{Avg.}} \\
\midrule

\rowcolor{groupgray}
\multicolumn{10}{l}{\textit{Models with 1.7--3B Parameters}} \\

Qwen3-1.7B (Nothinking)
& \no
& 10.42 & 8.33 & 6.67 & 3.33 & 5.00 & 10.00 & 0.00 & 6.25 \\

Qwen3-1.7B (Thinking)
& \no
& 37.50 & 33.33 & 34.58
& 19.58 & 15.00 & 17.25
& 0.00 & 22.46 \\

Distillation-1.7B
& \no
& 17.08 & 15.42 & 14.58 & 10.83 & 10.00 & 14.50
& \second{0.52} & 11.85 \\

OPD-1.7B
& \no
& 42.50 & 38.33 & 37.08 & 22.08 & \second{16.00} & 17.25
& \second{0.52} & 24.82 \\

Tool-Distillation-1.7B
& \yes
& 25.83 & 18.75 & 17.08 & 16.25 & 12.00 & 15.75
& \second{0.52} & 15.17 \\

Tool-OPD-1.7B  
& \yes
& 37.08 & 34.17 & 32.92 & 19.17 & 14.00 & 16.50
& \second{0.52} & 22.05 \\

SOD-1.7B
& \yes
& \best{47.08} & \second{40.42} & \second{40.83}
& \second{23.33} & \second{16.00} & \second{20.25}
& \second{0.52} & \second{26.92} \\

Tool-Star-3B
& \yes
& 17.92 & 12.50 & 13.75 & 10.42 & 6.00 & 10.50
& \second{0.52} & 10.23 \\

ARPO-3B
& \yes
& 18.75 & 19.17 & 23.33 & 10.83 & 5.00 & 11.75
& \best{2.08} & 12.99 \\

\rowcolor{oursblue}
\textbf{RIVET-1.7B}
& \yes
& \second{46.25} & \best{42.08} & \best{43.75} & \best{25.42}
& \best{19.00} & \best{20.75} & \second{0.52} & \best{28.25} \\

\rowcolor{expertgray}
\quad + External Experts
& \yes
& 55.83 & 50.42 & 52.92 & 33.75 & 27.00 & 31.50 & 2.60 & 36.29 \\

\midrule
\rowcolor{groupgray}
\multicolumn{10}{l}{\textit{Models with 4--32B Parameters}} \\

Qwen3-4B-Instruct
& \no
& 56.25 & 45.42 & 52.50 & 29.17 & 28.00 & 35.25 & 0.52 & 35.30 \\

Qwen3.5-4B (Nothinking)
& \no
& \second{67.92} & 48.75 & 56.25 & 38.33 & \best{35.00}
& 33.00 & 0.52 & 39.97 \\

Distillation-4B
& \no
& 61.67 & 49.58 & 56.25 & 33.75 & 29.00 & \second{35.50}
& 0.52 & 38.04 \\

Qwen3-14B (Thinking)
& \no
& \second{67.92} & \second{61.25} & \second{60.00}
& 39.58 & \second{34.00} & 31.75 & 0.00 & \second{42.07} \\

DPSK-14B
& \no
& 65.83 & 47.08 & 56.67 & 30.00 & 32.00 & 29.25 & 0.00 & 37.26 \\

Tool-Distillation-4B
& \yes
& 62.08 & 52.92 & 57.50 & 36.25 & 31.00 & 35.25
& 0.52 & 39.36 \\

SOD-Teacher-4B
& \yes
& 67.08 & 58.33 & 59.58 & 37.92 & \best{35.00} & 33.50
& 1.56 & 41.85 \\

TRICE-4B
& \yes
& 60.00 & 58.33 & 56.25 & 40.42 & 27.00 & 32.25
& 1.56 & 39.40 \\

DemyAgent-4B
& \yes
& 66.25 & \best{62.50} & 58.33 & \second{40.83}
& 29.00 & \second{35.50} & 1.56 & 42.00 \\

Agent Distillation-7B
& \yes
& 16.25 & 12.92 & 12.08 & 11.25 & 10.00 & 13.75
& 0.00 & 10.89 \\

ToRL-7B
& \yes
& 40.00 & 27.92 & 26.67 & 17.08 & 11.00 & 17.75
& 0.00 & 20.06 \\

ARPO-8B
& \yes
& 30.42 & 23.75 & 19.17 & 20.83 & 10.00 & 13.25 & 0.00 & 16.77 \\

AEPO-8B
& \yes
& 40.00 & 21.25 & 19.58 & 17.92 & 12.00 & 18.25 & 1.04 & 18.58 \\

ReTool-32B
& \yes
& 58.33 & 45.42 & 48.75 & 37.50 & 23.00 & 26.00
& \best{3.13} & 34.59 \\

\rowcolor{oursblue}
\textbf{RIVET-4B}
& \yes
& \best{68.75} & \second{61.25} & \best{61.67} & \best{42.08}
& \best{35.00} & \best{37.75} & \second{2.60} & \best{44.16} \\

\rowcolor{expertgray}
\quad + External Experts
& \yes
& 70.83 & 62.92 & 63.33 & 44.58 & 39.00 & 39.25 & 3.65 & 46.22 \\

\bottomrule
\end{tabular}
}
\end{threeparttable}
\end{table*}

\begin{table}[t]
  \centering
  \caption{Results on GPQA-Diamond. SOD-Teacher-4B is SOD's GRPO-trained teacher; expert-assisted rows are references excluded from the ranking.}
  \label{tab:gpqa-results}
  \begin{threeparttable}
  \renewcommand{\resultfont}{\scriptsize}
    \small
    \setlength{\tabcolsep}{5pt}
    \renewcommand{\arraystretch}{1.08}
    \scalebox{1}{%
    \begin{tabular}{lc>{\resultfont}S[table-format=2.2]}
      \toprule
      Method & Tool & \multicolumn{1}{c}{\makecell{GPQA\\Diamond}} \\
      \midrule

      \rowcolor{groupgray}
      \multicolumn{3}{l}{\textit{Models with 1.7--3B Parameters}} \\
      Qwen3-1.7B (Nothinking) & \no & 27.78 \\
      Qwen3-1.7B (Thinking) & \no & 36.87 \\
      Distillation-1.7B & \no & 30.81 \\
      OPD-1.7B & \no & \second{39.90} \\
      Tool-Distillation-1.7B  & \yes & 35.86 \\
      Tool-OPD-1.7B  & \yes & 37.37 \\
      SOD-1.7B  & \yes & \second{39.90} \\
      Tool-Star-3B & \yes & 27.78 \\
      ARPO-3B & \yes & 22.73 \\
      \rowcolor{oursblue}
      \textbf{RIVET-1.7B} & \yes & \best{44.95} \\
      \rowcolor{expertgray}
      \quad + External Experts & \yes & 54.04 \\

      \midrule
      \rowcolor{groupgray}
      \multicolumn{3}{l}{\textit{Models with 4--32B Parameters}} \\
      Qwen3-4B-Instruct & \no & 43.94 \\
      Distillation-4B & \no & 42.42 \\
    
      Qwen3-14B (Thinking) & \no & \second{59.09} \\
      DPSK-14B & \no & 55.05 \\
      Tool-Distillation-4B  & \yes & 43.94 \\
      
      SOD-Teacher-4B  & \yes & 57.07 \\
      TRICE-4B & \yes & 46.46 \\
      DemyAgent-4B & \yes & 54.55 \\
      ARPO-8B & \yes & 55.56 \\
      AEPO-8B & \yes & \best{61.62} \\
      ReTool-32B & \yes & 46.97 \\
      \rowcolor{oursblue}
      \textbf{RIVET-4B} & \yes & \best{61.62} \\
      \rowcolor{expertgray}
      \quad + External Experts & \yes & 67.68 \\
      \bottomrule
    \end{tabular}
    }
  \end{threeparttable}
\end{table}

\subsection{Experiment Setup}
\paragraph{Datasets and Metrics.}
We train on the ToRL dataset~\citep{li2025torlscalingtoolintegratedrl}.
Following~\citet{cheng2026teachingthinkingmodelsreason,yu2025demystifyingreinforcementlearningagentic}, we evaluate seven mathematics benchmarks: AIME2024--2026~\citep{aime24,aime25,aime26}, HMMT2025~\citep{dekoninck2026benchmarksmatharenaevaluationplatform}, BeyondAIME~\citep{seed2025seed15thinkingadvancingsuperbreasoning}, IMO-AnswerBench~\citep{luong2025robustmathematicalreasoning}, and APEX2025~\citep{dekoninck2026benchmarksmatharenaevaluationplatform}. We additionally use GPQA-Diamond to test whether the learned capability generalizes beyond competition mathematics to scientific reasoning~\citep{rein2023gpqagraduatelevelgoogleproofqa}. Accuracy is the sole task-performance metric. We report \emph{Mean@8} for AIME2024--2026 and HMMT2025, and \emph{Mean@16} for APEX2025.
\paragraph{Baselines.} We group baselines by tool availability at inference. \textbf{Tool-free reasoning models} include Qwen3-1.7B, Qwen3-4B-Instruct, and Qwen3-14B~\citep{yang2025qwen3technicalreport}; Qwen3.5-4B~\citep{qwen3.5}; and DeepSeek-R1-Distill-Qwen-14B (DPSK-14B)~\citep{Guo_2025}. We also construct Distillation-1.7B and Distillation-4B from correct Qwen3.5-9B solutions to RIVET's training problems, and train OPD-1.7B through on-policy distillation~\citep{agarwal2024onpolicydistillationlanguagemodels} from Qwen3-8B to Qwen3-1.7B in thinking mode without tools. \textbf{Tool-integrated reasoning agents} use execution or retrieval feedback. They include Tool-Star-3B~\citep{dong2025toolstarempoweringllmbrainedmultitool}, TRICE-4B~\citep{cheng2026teachingthinkingmodelsreason}, DemyAgent-4B~\citep{yu2025demystifyingreinforcementlearningagentic}, ToRL-7B~\citep{li2025torlscalingtoolintegratedrl}, ARPO-3B and ARPO-8B~\citep{dong2025agenticreinforcedpolicyoptimization}, AEPO-8B~\citep{dong2025agenticentropybalancedpolicyoptimization}, Agent Distillation-7B~\citep{kang2025distillingllmagentsmall}, SOD-1.7B and its GRPO-trained teacher (SOD-Teacher-4B)~\citep{zhong2026sodstepwiseonpolicydistillation}, and ReTool-32B~\citep{feng2025retoolreinforcementlearningstrategic}. We additionally construct Tool-Distillation-1.7B and Tool-Distillation-4B from verified trajectories independently generated by Qwen3.5-9B with access to a Python tool on the same training problems. We also train Tool-OPD-1.7B through on-policy distillation in non-thinking mode with access to a Python tool. RIVET's internalized mode uses local Python with no external LLM; the $+$ External Experts rows report assisted-inference references. Appendix~\ref{apx:baselines} details the baseline configurations.

\paragraph{Training Details.} We train Qwen3-1.7B and Qwen3-4B-Instruct using veRL~\citep{sheng2024hybridflow} on NVIDIA A800 GPUs. Stage~I uses three frozen experts: Qwen3.5-9B, Qwen3-8B, and Qwen2.5-Coder-7B-Instruct.  Stage~II uses two SFT epochs. The training setup and hyperparameters for both stages are provided in Appendix~\ref{app:stage2-training-config}.

\subsection{Main Results}
\paragraph{Internalized performance on mathematics and scientific reasoning.}
RIVET-1.7B and RIVET-4B lead their respective model-size groups on competition mathematics, with average accuracies of 28.25\% and 44.16\% (Table~\ref{tab:main-results}). RIVET-4B also scores higher than substantially larger models, including ReTool-32B. On GPQA-Diamond (Table~\ref{tab:gpqa-results}), RIVET-1.7B ranks first in the smaller-model group, while RIVET-4B ties AEPO-8B for first place in the larger-model group. Both RIVET models produce these results with local Python and no external LLM, providing evidence that the learned capability generalizes to scientific reasoning.

\paragraph{Improvements over offline and on-policy distillation.}
At both model sizes, \textsc{Rivet} scores higher than the offline Distillation and Tool-Distillation baselines on the mathematics average and GPQA-Diamond. At 1.7B, it also scores higher than OPD, Tool-OPD, and the SOD student on both measures. SOD-1.7B leads on AIME 2024, while RIVET-1.7B has the higher mathematics average. These results compare the complete RIVET pipeline with the evaluated distillation configurations; the ablations below examine Stage-II training and the routing policies that produce its initialization and trajectories.

\begin{table}[t]
\centering
\caption{Stage~II improves RIVET-4B's accuracy after expert removal. All rows use local Python without external LLMs.}
\label{tab:expert-ablation}
\scriptsize
\setlength{\tabcolsep}{1.5pt}
\renewcommand{\arraystretch}{1.12}
\resizebox{\columnwidth}{!}{%
\begin{tabular}{@{}lcccccccc@{}}
\toprule
Method
& \makecell{AIME\\2024}
& \makecell{AIME\\2025}
& \makecell{AIME\\2026}
& \makecell{HMMT\\2025}
& \makecell{Beyond\\AIME}
& \makecell{IMO\\Answer}
& \makecell{APEX\\2025}
& Avg. \\
\midrule
\textbf{RIVET-4B}
& \textbf{68.75} & \textbf{61.25} & \textbf{61.67}
& \textbf{42.08} & \textbf{35.00} & 37.75
& \textbf{2.60} & \textbf{44.16} \\
\makecell[l]{Stage I Remove Experts}
& 61.25 & 47.08 & 55.42 & 32.92
& 30.00 & 36.00 & 1.04 & 37.67 \\
\makecell[l]{Training w/o Experts}
& 59.58 & 50.42 & 54.58 & 33.75
& 28.00 & \textbf{39.75} & 1.04 & 38.16 \\
\bottomrule
\end{tabular}%
}
\end{table}

\subsection{Ablation Study}

\paragraph{Expert removal exposes a gap that Stage~II reduces.} The Stage-I controller reaches $46.22\%$ with experts (Table~\ref{tab:main-results}) and $37.67\%$ when they are disabled (Table~\ref{tab:expert-ablation}), an $8.55$-point gap. Stage~II raises accuracy under the latter deployment interface to $44.16\%$, a $6.49$-point improvement without external LLMs. Supervised internalization thus improves performance beyond the expert-span training already included in collaborative RL. A variant trained without external experts throughout the pipeline obtains $38.16\%$, or $6.00$ points below RIVET, with the same self-reasoning and local-Python interface. The next ablation examines how trajectory training and the Stage-II objective contribute to the improvement.

\begin{table}[t]
\centering
\caption{Moderate format weighting improves on ordinary Stage-II SFT ($\lambda_{\mathrm{fmt}}=0$) for Qwen3-4B-Instruct.}
\label{tab:format-loss-ablation}
\scriptsize
\setlength{\tabcolsep}{2.5pt}
\renewcommand{\arraystretch}{1.05}
\scalebox{1.0}{%
\begin{tabular}{@{}lcccccccc@{}}
\toprule
$\lambda_{\mathrm{fmt}}$
& \makecell{AIME\\24}
& \makecell{AIME\\25}
& \makecell{AIME\\26}
& \makecell{HMMT\\25}
& \makecell{Beyond\\AIME}
& \makecell{IMO\\Answer}
& \makecell{APEX\\25}
& Avg. \\
\midrule
$1.0$ & 66.25 & 57.08 & 60.42 & 39.17 & \textbf{37.00} & 36.50 & 2.08 & 42.64 \\
$0.5$ & \textbf{68.75} & \textbf{61.25} & \textbf{61.67} & \textbf{42.08} & 35.00 & \textbf{37.75} & \textbf{2.60} & \textbf{44.16} \\
$0.0$ & 64.58 & 55.83 & 59.17 & 37.92 & 33.00 & 34.25 & 1.56 & 40.90 \\
\bottomrule
\end{tabular}
}
\end{table}

\begin{table*}[t]
\centering
\caption{Effect of Stage-I expert selection on the complete two-stage pipeline with Qwen3-4B-Instruct. Stage-I performance is evaluated with external experts, whereas Stage-II performance is evaluated without them.}
\label{tab:routing-ablation}
\scriptsize
\setlength{\tabcolsep}{3.2pt}
\renewcommand{\arraystretch}{1.10}
\begin{adjustbox}{max width=\textwidth}
\begin{tabular}{@{}llcccccccc@{}}
\toprule
Method
& Evaluation
& \makecell{AIME\\2024}
& \makecell{AIME\\2025}
& \makecell{AIME\\2026}
& \makecell{HMMT\\2025}
& \makecell{Beyond\\AIME}
& \makecell{IMO\\Answer}
& \makecell{APEX\\2025}
& Avg. \\
\midrule
\rowcolor{oursblue}
\textbf{RIVET-4B}
& Stage~I (collaborative)
& 70.83 & \textbf{62.92} & \textbf{63.33} & \textbf{44.58}
& \textbf{39.00} & \textbf{39.25} & \textbf{3.65} & \textbf{46.22} \\
\rowcolor{oursblue}
& Stage~II (internalized)
& 68.75 & \textbf{61.25} & \textbf{61.67} & \textbf{42.08}
& 35.00 & \textbf{37.75} & \textbf{2.60} & \textbf{44.16} \\
\addlinespace[2pt]
Qwen3.5-9B Only
& Stage~I (collaborative)
& \textbf{71.25} & 59.58 & 60.42 & 41.25
& \textbf{39.00} & 36.75 & 2.60 & 44.41 \\
& Stage~II (internalized)
& \textbf{69.58} & 57.08 & 57.92 & 37.92
& \textbf{36.00} & 34.00 & 2.08 & 42.08 \\
\addlinespace[2pt]
Random Selection
& Stage~I (collaborative)
& 63.75 & 56.25 & 58.75 & 39.58
& 34.00 & 35.50 & 3.13 & 41.57 \\
& Stage~II (internalized)
& 61.25 & 54.58 & 55.83 & 36.25
& 30.00 & 33.25 & 2.08 & 39.03 \\
\bottomrule
\end{tabular}
\end{adjustbox}
\end{table*}

\paragraph{Trajectory SFT and format weighting both improve post-removal accuracy.}
Ordinary SFT on verified trajectories ($\lambda_{\mathrm{fmt}}=0$) raises accuracy from $37.67\%$ to $40.90\%$, a $3.23$-point gain (Tables~\ref{tab:expert-ablation} and~\ref{tab:format-loss-ablation}). Adding format weight $\lambda_{\mathrm{fmt}}=0.5$ raises it to $44.16\%$, a further $3.26$ points. Stronger weighting ($\lambda_{\mathrm{fmt}}=1.0$) obtains $42.64\%$. These sequential gains show that ordinary cross-entropy training on complete successful interactions improves accuracy after expert removal, with additional gains from emphasizing tool-call structure. We use $\lambda_{\mathrm{fmt}}=0.5$ in the remaining experiments.

\begin{figure}[t]
    \centering
    \includegraphics[width=0.90\columnwidth]{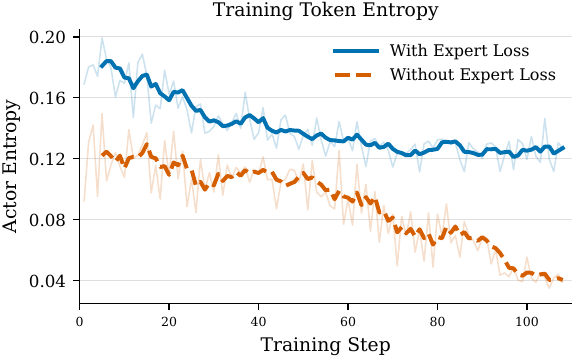}
    \caption{Controller entropy is higher with expert-span updates during the early Stage~I steps shown. Bold curves show the five-step moving average, while faint curves show per-step values.}
    \label{fig:expert_loss_entropy}
\end{figure}

\paragraph{Expert-span updates maintain higher controller entropy in the observed run.}
Figure~\ref{fig:expert_loss_entropy} compares the full objective in Eq.~\eqref{eq:stage1_loss} with a controller-token-only variant, $m^{\mathrm{train}}=m^{\mathrm{ctrl}}$. Both curves report controller-policy entropy. Over the displayed interval, the full objective maintains higher entropy, while entropy in the controller-only variant continues to decline. The controller-only run did not complete Stage~I; no final Stage-I or Stage-II accuracy is reported. This comparison characterizes an optimization difference associated with expert-span updates, without isolating their effect on final task accuracy.

\begin{figure*}[t]
    \centering
    \includegraphics[width=\textwidth]
    {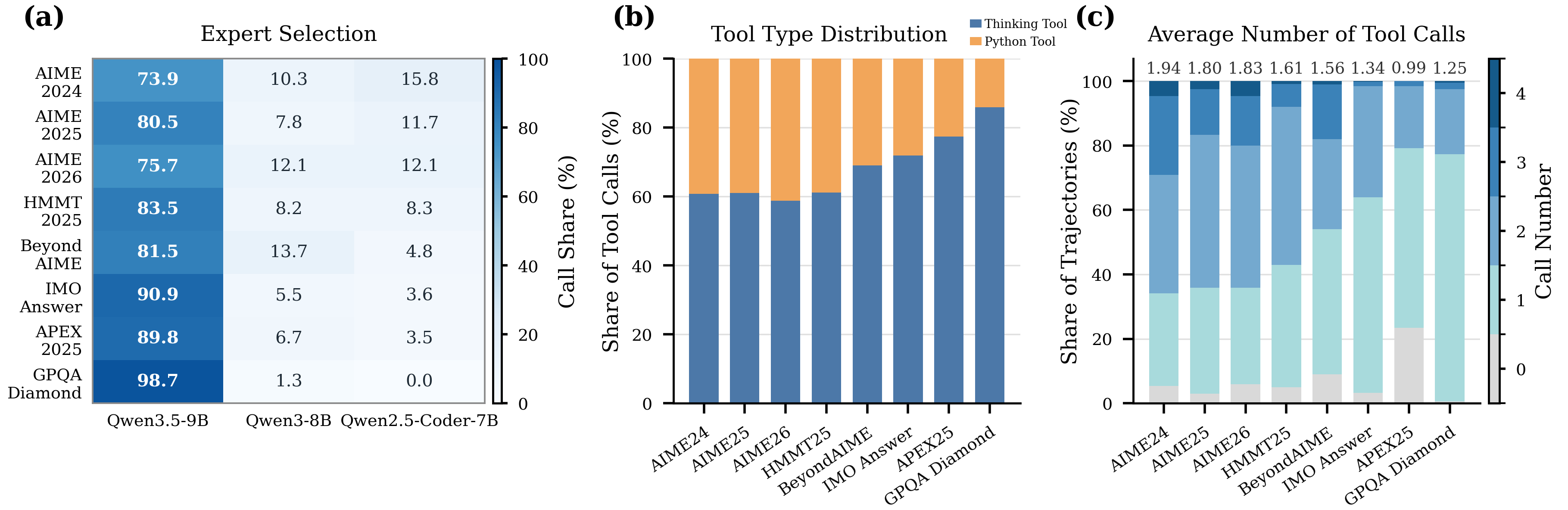}
    \caption{Routing behavior of the Stage-I controller when
    interacting with external experts. (a) Distribution of
    external expert calls across the three heterogeneous experts.
    (b) Distribution of thinking and Python tool calls.
    (c) Number of tool calls per trajectory; values above the bars
    report the mean number of calls.}
    \label{fig:expert-routing-analysis}
\end{figure*}

\paragraph{Collaboration policy affects performance after internalization.}
We compare RIVET-4B with variants that use only Qwen3.5-9B or select experts randomly (Table~\ref{tab:routing-ablation}). Each starts Stage~II from its own Stage-I checkpoint and trains on 12,000 verified trajectories with the same SFT hyperparameters. Learned routing gives the highest collaborative average and leads to $44.16\%$ after internalization, compared with $42.08\%$ for Qwen3.5-9B-only routing and $39.03\%$ for random selection. The advantage persists after expert removal in this comparison of complete two-stage variants, where routing affects both the Stage-II initialization and its training trajectories. Appendix~\ref{app:routing-ablation-setup} details the protocol.

\subsection{Analysis Experiments}

We first examine which successes survive expert removal, then characterize the expert content elicited by the collaboration policy through routing distributions and call-budget analysis.

\begin{table}[t]
\centering
\caption{Retention and internalization rates of RIVET-4B.}
\label{tab:internalization-analysis}
\small
\setlength{\tabcolsep}{2pt}
\renewcommand{\arraystretch}{1.10}
\scalebox{0.90}{%
\begin{tabular}{@{}lcc@{}}
\toprule
Benchmark
& \makecell{Retention Rate}
& \makecell{Internalization Rate} \\
\midrule
AIME Family   & 81.17 & 53.01 \\
HMMT 2025     & 74.68 & 46.43 \\
BeyondAIME    & 80.00 & 50.00 \\
GPQA-Diamond  & 84.55 & 55.56 \\
\bottomrule
\end{tabular}
}
\end{table}

\paragraph{Internalization preserves existing successes and recovers assisted ones.} Table~\ref{tab:internalization-analysis} compares the same three model states introduced above: Stage~I with experts, Stage~I without experts, and the internalized Stage-II controller. Retention is the percentage of cases solved without experts before Stage~II that remain correct afterward. Internalization is the percentage solved by Stage~II among cases that Stage~I solves with experts but fails without them. Across the reported task groups, RIVET-4B retains 74.68--84.55\% of independent successes and recovers 46.43--55.56\% of expert-rescued cases. These rates measure instance-level preservation and recovery on different conditioning sets, rather than the aggregate accuracy differences in the ablation study.

\paragraph{Expert and tool choices vary across tasks.} The Stage-I RIVET-4B controller exhibits different routing distributions across benchmarks in collaborative mode. Figure~\ref{fig:expert-routing-analysis}(a) shows a higher share of code-expert calls on AIME tasks, greater use of Qwen3-8B on BeyondAIME, and near-exclusive use of Qwen3.5-9B on GPQA-Diamond. Figure~\ref{fig:expert-routing-analysis}(b) shows a larger Python share on competition-mathematics tasks than on the scientific-reasoning benchmark. Figure~\ref{fig:expert-routing-analysis}(c) shows task-dependent call counts, averaging fewer than two per trajectory on each benchmark despite a maximum budget of four. These distributions characterize the expert reasoning and code encountered during collaboration; the routing ablation measures performance of the resulting two-stage variants.

\begin{figure}[t]
    \centering
    \includegraphics[width=0.90\columnwidth]{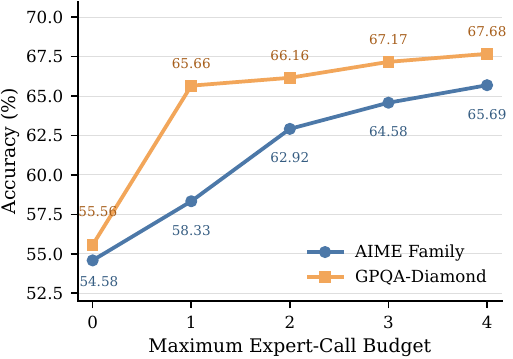}
    \caption{Accuracy of RIVET-4B under different maximum external expert call budgets. }
    \label{fig:expert-call-budget}
\end{figure}

\paragraph{The marginal benefit of expert calls depends on the task.} We vary the maximum number of external-expert calls while holding all other inference settings fixed. Figure~\ref{fig:expert-call-budget} shows that allowing one call produces most of the gain on GPQA-Diamond, with smaller improvements at higher budgets. The AIME family continues to benefit from additional calls across the tested budgets. Thus, expert access improves accuracy in both task groups, but the benefit of a larger call budget differs between them. Together with the routing distributions, this analysis characterizes how the controller acquires assistance before internalization.

\section{Conclusion}
RIVET trains a controller on both the decisions that organize expert assistance and the reasoning and code experts provide. Expert-augmented GRPO updates controller and expert spans using a shared outcome signal; verified trajectory internalization further consolidates their sequential relationship through format-aware supervised training. RIVET-4B reaches $44.16\%$ mathematics accuracy with local Python and no external LLMs, a $6.49$-point improvement over its Stage-I checkpoint under the same deployment interface. Ablations show gains from ordinary trajectory supervision and additional format weighting, while the routing comparison measures the effect of collaboration policy on the complete two-stage pipeline. GPQA-Diamond provides evidence of generalization to scientific reasoning. These results support training on complete collaborations to extend a controller from coordinating expert assistance to generating reasoning and code itself.

\section*{Limitations}
Our evaluation covers two Qwen-based controller scales and three frozen experts. The deployment setting is external-LLM-free but retains local Python, so the results concern tool-integrated reasoning. Expert calls add training cost; we report call counts but do not quantify total token cost or the deployment volume at which internalization becomes cost-effective. The expert-span ablation characterizes optimization behavior in one run, while the strong dependence on Stage-II format weighting leaves the relative contributions of semantic learning and interface supervision unresolved. Accuracy is the sole task-performance metric. Establishing broader generality requires additional model families, expert pools, domains, and random seeds.

\clearpage


\bibliography{custom}

\clearpage
\appendix

\section{Appendix}

\subsection{Controller Prompt and Tool Definitions}
\label{app:controller-prompt}

Table~\ref{tab:controller-input-prompt} presents the controller input template from a logged Stage-I Qwen3-4B-Instruct rollout. The native \texttt{tools} argument supplies the two function definitions in Table~\ref{tab:controller-tool-definitions}, which the chat template inserts into the system message. For presentation, \texttt{<TOOL\_SCHEMAS>} marks their location and \texttt{<PROBLEM>} replaces only the task-specific question. Prompt wording, parameter descriptions, and required fields are otherwise preserved.

\begingroup
\definecolor{rivetpromptheader}{RGB}{232,239,247}
\definecolor{rivetpromptink}{RGB}{37,61,85}
\lstdefinestyle{rivetprompt}{
  language={},
  basicstyle=\ttfamily\fontsize{8.1}{9.5}\selectfont,
  columns=fullflexible,
  keepspaces=true,
  breaklines=true,
  breakatwhitespace=true,
  breakautoindent=false,
  breakindent=0pt,
  showstringspaces=false,
  aboveskip=3pt,
  belowskip=3pt,
  tabsize=2
}

\begin{table*}[p]
\centering
\caption{Controller input template for Stage-I collaboration. Shaded headings identify message roles; tool schemas are given separately in Table~\ref{tab:controller-tool-definitions}.}
\label{tab:controller-input-prompt}
\setlength{\tabcolsep}{8pt}
\renewcommand{\arraystretch}{1.12}
\begin{tabular}{p{\textwidth}}
\toprule
\rowcolor{rivetpromptheader}
\strut\hspace{6pt}\textbf{\textcolor{rivetpromptink}{System message: native tool-calling interface}} \\
\begin{minipage}{\dimexpr\linewidth-12pt\relax}
\lstset{style=rivetprompt}
\begin{lstlisting}
# Tools

You may call one or more functions to assist with the user query.

You are provided with function signatures within <tools></tools> XML tags:
<tools>
<TOOL_SCHEMAS>
</tools>

For each function call, return a json object with function name and arguments within <tool_call></tool_call> XML tags:
<tool_call>
{"name": <function-name>, "arguments": <args-json-object>}
</tool_call>
\end{lstlisting}
\end{minipage} \\
\midrule
\rowcolor{rivetpromptheader}
\strut\hspace{6pt}\textbf{\textcolor{rivetpromptink}{User message: collaboration instructions and problem}} \\
\begin{minipage}{\dimexpr\linewidth-12pt\relax}
\lstset{style=rivetprompt}
\begin{lstlisting}
You are a tool-using math controller.

The external expert tools are available to improve correctness. Use them actively but efficiently:
- For non-trivial problems, prefer at least one useful expert/tool step when it can improve correctness.
- Use `think` for planning, derivation, critique, or deciding the next step. Use `code_interpreter` when the next useful step is computation, symbolic manipulation, case checking, or verification.
- For multi-step or uncertain problems, use alternating tool evidence and reasoning such as `think -> code_interpreter -> think`, then finish once the answer is clear.
- Keep controller reasoning concise. Use expert/tool evidence to avoid long self-derivations.
- After receiving expert/tool evidence, do not repeat a full derivation. Summarize only the key evidence and finish with the final boxed answer once confident.
- When calling `code_interpreter` with an external expert model, set `code` to an empty string and put the request in `instruction`. Use `model="self"` only for simple direct code execution or small fixes.
- Do not add extra tool calls after the answer is already determined. The final line must be exactly: Answer: \boxed{answer}

Available strong expert models:
- Qwen3.5-9B
- Qwen2.5-Coder-7B-Instruct
- Qwen3-8B

Use the tool-call JSON format shown in the system Tools section. The final line must be exactly: Answer: \boxed{answer}

# Problem
Solve the following math problem step by step. The last line of your response should be exactly of the form Answer: \boxed{answer}, where answer is the answer to the problem.

<PROBLEM>

Remember to put your final answer on its own line using exactly: Answer: \boxed{answer}.
\end{lstlisting}
\end{minipage} \\
\midrule
\strut\hspace{6pt}\textbf{Generation prefix:}\quad\texttt{assistant} \\
\bottomrule
\end{tabular}
\end{table*}

\begin{table*}[p]
\centering
\caption{Complete native tool definitions supplied to the Stage-I controller. These two objects form the \texttt{tools} list and are serialized inside the \texttt{<tools>} block in Table~\ref{tab:controller-input-prompt}. JSON whitespace is reformatted for readability.}
\label{tab:controller-tool-definitions}
\setlength{\tabcolsep}{7pt}
\renewcommand{\arraystretch}{1.12}
\begin{tabular}{p{\dimexpr.5\textwidth-7pt\relax}p{\dimexpr.5\textwidth-7pt\relax}}
\toprule
\rowcolor{rivetpromptheader}
\strut\textbf{\textcolor{rivetpromptink}{Reasoning tool: \texttt{think}}}
& \strut\textbf{\textcolor{rivetpromptink}{Python tool: \texttt{code\_interpreter}}} \\
\begin{minipage}[t]{\linewidth}
\lstset{style=rivetprompt,basicstyle=\ttfamily\fontsize{8}{9.3}\selectfont}
\begin{lstlisting}
{
  "type": "function",
  "function": {
    "name": "think",
    "description": "Ask a strong external expert model to produce a concise reasoning/planning step. These experts are stronger specialist solvers than the controller; call this tool often for non-trivial planning, derivation, critique, or answer checks, but never call think twice in a row. The returned text is appended to the trajectory for the controller to use next. Qwen2.5-Coder-7B-Instruct is especially strong at code, while Qwen3.5-9B is especially strong at reasoning and planning.",
    "parameters": {
      "type": "object",
      "properties": {
        "model": {
          "type": "string",
          "enum": [
            "Qwen3.5-9B",
            "Qwen2.5-Coder-7B-Instruct",
            "Qwen3-8B"
          ],
          "description": "Strong external expert model to ask for the reasoning/planning step."
        },
        "instruction": {
          "type": "string",
          "description": "Optional short instruction for what this expert should think about."
        }
      },
      "required": [
        "model"
      ]
    }
  }
}
\end{lstlisting}
\end{minipage}
&
\begin{minipage}[t]{\linewidth}
\lstset{style=rivetprompt,basicstyle=\ttfamily\fontsize{8}{9.3}\selectfont}
\begin{lstlisting}
{
  "type": "function",
  "function": {
    "name": "code_interpreter",
    "description": "Run a Python code-interpreter step. Use model='self' when the controller writes executable code directly; prefer an external expert model name for non-trivial code generation, checks, or symbolic work. Code is executed in an isolated sandbox.",
    "parameters": {
      "type": "object",
      "properties": {
        "model": {
          "type": "string",
          "enum": [
            "self",
            "Qwen3.5-9B",
            "Qwen2.5-Coder-7B-Instruct",
            "Qwen3-8B"
          ],
          "description": "Prefer selecting a strong external expert model to generate Python code. Use self only for simple direct execution or small fixes."
        },
        "instruction": {
          "type": "string",
          "description": "Optional short instruction for the Python step."
        },
        "code": {
          "description": "Required field for python tool calls. When model is self, provide executable Python code for one isolated call. The last bare expression is automatically displayed like Jupyter, and print(...) output is also returned. Do not rely on variables from previous calls. When model is an external expert, set code to an empty string; the expert will generate the code.",
          "type": "string"
        }
      },
      "required": [
        "model",
        "code"
      ]
    }
  }
}
\end{lstlisting}
\end{minipage} \\
\bottomrule
\end{tabular}
\end{table*}
\endgroup

\subsection{Baseline Details}
\label{apx:baselines}

\paragraph{Tool-free reasoning models.} These models are evaluated without external tools. We enable Qwen3-14B's native thinking mode and evaluate Qwen3-1.7B in both thinking and non-thinking settings. Qwen3-4B-Instruct and DeepSeek-R1-Distill-Qwen-14B use their official chat templates with external tools disabled.

\paragraph{Distillation baselines.} We train two supervised-distillation baselines on the same problems as RIVET. For each problem, Qwen3.5-9B generates eight candidate responses; those with correct final answers are retained as teacher supervision. Fine-tuning Qwen3-1.7B and Qwen3-4B-Instruct on these trajectories yields Distillation-1.7B and Distillation-4B, respectively. Both are evaluated without the teacher or external tools.

\paragraph{On-policy distillation.} We construct two on-policy distillation baselines using Qwen3-8B as the teacher and Qwen3-1.7B as the student. Both variants use the same training problems as RIVET and a training batch size of 32. For OPD-1.7B, the teacher provides supervision in thinking mode without tools, and the student is trained and evaluated in the same setting. For Tool-OPD-1.7B, both teacher and student operate in non-thinking mode with access to a Python tool. At inference, OPD-1.7B reasons without tools, whereas Tool-OPD-1.7B can invoke the Python tool; neither student accesses the teacher or any other external LLM. We do not construct an OPD baseline for Qwen3-4B-Instruct because Qwen3-8B in non-thinking mode underperforms this student in our setting.

\paragraph{Tool-augmented distillation baselines.} We additionally construct Tool-Distillation-1.7B and Tool-Distillation-4B to control for access to local Python. For each training problem, we provide Qwen3.5-9B with a Python tool definition through its native tool-calling interface and sample eight candidate trajectories. The teacher generates structured Python tool calls and receives sandbox execution results as feedback for subsequent reasoning. We retain complete trajectories with correct final answers and valid Python execution when applicable, preserving the native tool-call and response structure when fine-tuning Qwen3-1.7B and Qwen3-4B-Instruct. Teacher reasoning, tool calls, code, and final answers are prediction targets, whereas problem prompts and execution outputs are masked from the loss. At inference, both students use the same local Python interface as internalized RIVET, without access to the teacher or any other external LLM.

\paragraph{Tool-integrated reasoning agents.} These agents incorporate observations from external tools into subsequent reasoning. We evaluate released checkpoints where available. ReTool interleaves reasoning with multi-turn code execution, using cold-start fine-tuning followed by RL; we evaluate ReTool-32B. TRICE combines tool-integrated supervised fine-tuning with RL to train interleaved reasoning and code execution; we evaluate TRICE-4B. Tool-Star combines synthesized tool-use trajectories with multi-tool self-critic RL; we evaluate Tool-Star-3B. DemyAgent trains on end-to-end tool-use trajectories with exploration-oriented agentic RL. We reproduce it with Qwen3-4B-Instruct as DemyAgent-4B. ToRL incorporates a code interpreter into RL rollouts to learn from execution feedback; we evaluate \texttt{GAIR/ToRL-7B} as ToRL-7B. ARPO trains multi-turn tool-use policies through agentic RL; we evaluate \texttt{Qwen2.5-3B-ARPO} and \texttt{Qwen3-8B-ARPO-DeepSearch} as ARPO-3B and ARPO-8B, respectively. AEPO uses entropy-balanced policy optimization to maintain exploration; we evaluate \texttt{Qwen3-8B-AEPO-DeepSearch} as AEPO-8B. Agent Distillation transfers teacher-agent behavior involving retrieval and code execution through supervised learning; we evaluate its released 7B checkpoint as Agent Distillation-7B. SOD introduces adaptive step-level weighting for on-policy distillation of tool-using agents; we evaluate its released 1.7B student as SOD-1.7B and its GRPO-trained 4B teacher as SOD-Teacher-4B. The latter is a teacher checkpoint, not a 4B student trained with SOD. The Agent Distillation and SOD checkpoints are evaluated without further fine-tuning.

\paragraph{Evaluation protocol.} We evaluate all baselines on NVIDIA A800 GPUs. Unless an official method requires a different configuration, we use temperature 0.6, top-p 0.95, and a maximum output length of 16{,}384 tokens. Tool-free baselines use their official chat templates with external tool interfaces disabled. For tool-integrated baselines, we preserve the released tool specification, system prompt, and interaction protocol, changing only the benchmark-specific problem statement and required answer format. We report \emph{Mean@8} on AIME 2024, AIME 2025, AIME 2026, and HMMT 2025; \emph{Mean@16} on APEX 2025; and single-run accuracy on BeyondAIME, IMO-AnswerBench, and GPQA-Diamond.

\subsection{Training Configuration}
\label{app:stage2-training-config}

\paragraph{Training setup.} We train Qwen3-1.7B and Qwen3-4B-Instruct using veRL~\citep{sheng2024hybridflow} on NVIDIA A800 GPUs. Stage~I uses three frozen experts: Qwen3.5-9B, Qwen3-8B, and Qwen2.5-Coder-7B-Instruct. Table~\ref{tab:stage2-training-config} summarizes the training settings for Stage-I GRPO and Stage-II supervised fine-tuning.

\begin{table}[t]
\centering
\caption{Training settings for the two stages of RIVET.}
\label{tab:stage2-training-config}
\small
\setlength{\tabcolsep}{6pt}
\renewcommand{\arraystretch}{1.08}
\begin{tabular}{@{}lc@{}}
\toprule
Hyperparameter & Value \\
\midrule
\multicolumn{2}{l}{\textit{Stage I: GRPO}} \\
Learning rate & $1\times10^{-6}$ \\
Training steps & 300 \\
Global batch size & 32 \\
Rollouts per prompt & 8 \\
Advantage stabilizer $\epsilon_A$ & $10^{-6}$ \\
PPO clip $\epsilon$ & $0.2$ \\
Dual-clip $c$ & $3$ \\
Maximum rollout length & 8k tokens \\
Maximum expert response length & 3k tokens \\
Maximum expert calls per rollout & 4 \\
Random seed & 66 \\
\midrule
\multicolumn{2}{l}{\textit{Stage II: SFT}} \\
Learning rate & $2\times10^{-6}$ \\
Global batch size & 128 \\
Training epochs & 2 \\
Maximum sequence length & 16,384 tokens \\
\bottomrule
\end{tabular}
\end{table}

\subsection{Details of Ablation Experiments}
\label{app:routing-ablation-setup}
\paragraph{Routing ablation setup.} All variants in Table~\ref{tab:routing-ablation} use exactly the same training problems as RIVET-4B, with expert selection set to learned routing, Qwen3.5-9B-only routing, or random selection. Their Stage-II SFT datasets are constructed using the same trajectory generation and filtering procedure described in Section~\ref{sec:stage2}: eight candidate trajectories are sampled per training problem, followed by correctness and validity filtering, deduplication, and manual review, with at most two retained trajectories per problem. Each variant uses 12,000 verified trajectories and the same SFT hyperparameters as RIVET-4B, including the batch size, learning rate, number of training epochs, and format-loss weight $\lambda_{\mathrm{fmt}}=0.5$. Each variant initializes Stage~II from its own trained Stage-I checkpoint. We train for two full SFT epochs and evaluate the final checkpoint after the second epoch. This comparison evaluates the complete two-stage variants, allowing both the Stage-II initialization and the training trajectories to depend on the Stage-I routing policy.

\subsection{Stage-II Trajectory Conversion Example}
\label{app:stage2-conversion-example}

\paragraph{Canonicalization and target ownership.} Stage~II preserves the native \texttt{<tool\_call>}--\texttt{<tool\_response>} serialization, selected expert, call order, code field, and returned content. Missing or free-form instructions are replaced by fixed tool-specific instructions. The retained expert reasoning and code are included as controller prediction targets in the supervised sequence, whereas actual Python execution results remain masked environment observations. Stage~I already trains on expert spans; the conversion here specifies their place in the complete Stage-II target sequence and subsequent internalized decoding.

\paragraph{Decoding and execution in internalized mode.} The Stage-II controller generates the tool-call structure, serialized role markers, and the reasoning and code formerly supplied by experts within the same decoding stream. Thus, \texttt{user <tool\_response>} does not indicate a new external message, and the retained \texttt{model} field does not dispatch a call to the named expert. Decoding continues through these spans and pauses for tool execution when a complete \texttt{<python>}--\texttt{</python>} block has been generated. The runtime executes the code in this block, rather than the empty \texttt{code} argument shown in the example, appends the actual result within \texttt{<output>}--\texttt{</output>}, and resumes controller generation from the updated context. This cycle continues until the final answer; no external LLM is invoked. Stage~I already includes expert-origin reasoning and code in the controller's training loss, and Stage~II further supervises these spans while assigning additional loss weight to tool-call format tokens.

\paragraph{Complete before--after example.} Table~\ref{tab:stage2-conversion-example} reproduces both sides of a verified conversion without abbreviation. Panel~(a) gives the original Stage-I collaborative trajectory, and Panel~(b) gives the corresponding Stage-II training sequence together with its target mask. The panels show the surface-form canonicalization and distinguish each span's original producer from its role in supervised training.

The before--after comparison makes both transformations explicit. The missing thinking instruction and problem-specific Python instruction in Panel~(a) are mapped to fixed tool-specific instructions in Panel~(b). Reasoning and code originally generated by Qwen3.5-9B are supervised controller targets in Panel~(b), while the sandbox output remains contextual input. This converted training sequence illustrates how the expert's derivation and verification program are used to train the controller; it is not a separately sampled Stage-II inference trajectory.

\begin{table*}[p]
\centering
\caption{A complete before--after example of Stage-II trajectory conversion. Panel~(a) shows the original Stage-I collaborative trajectory. Panel~(b) shows the corresponding Stage-II sequence, in which instructions are canonicalized and former expert outputs become controller targets. Both panels reproduce the complete problem, expert reasoning, Python program, execution result, and final response.}
\label{tab:stage2-conversion-example}
\scriptsize
\setlength{\tabcolsep}{3.5pt}
\renewcommand{\arraystretch}{1.08}
\begin{tabular}{@{}>{\raggedright\arraybackslash}p{0.07\textwidth}>{\raggedright\arraybackslash}p{0.14\textwidth}>{\raggedright\arraybackslash}p{0.74\textwidth}@{}}
\toprule
\multicolumn{3}{@{}l}{\textbf{(a) Original Stage-I collaborative trajectory (Part 1 of 2)}} \\
\addlinespace[2pt]
\textbf{Turn} & \textbf{Producer} & \textbf{Original serialized content} \\
\midrule

Input
& Problem prompt
& Solve the following math problem step by step. The last line of your response should be exactly of the form \texttt{Answer: \textbackslash boxed\{answer\}}, where \texttt{answer} is the answer to the problem. \newline
  Determine how many right triangles exist with integer leg lengths $a$ and $b$, where the hypotenuse is of length $b+1$ and $b<100$. \newline
  Remember to put your final answer on its own line using exactly: \texttt{Answer: \textbackslash boxed\{answer\}}. \\
\addlinespace[2pt]

Call 1
& Controller
& \cellcolor{blue!3}\texttt{<tool\_call>} \newline
  \texttt{\{"name": "think", "arguments": \{"model": "Qwen3.5-9B"\}\}} \newline
  \texttt{</tool\_call>} \\
\addlinespace[2pt]

Response 1
& Qwen3.5-9B
& \cellcolor{blue!3}\texttt{user <tool\_response>} \newline
  I need to find the number of right triangles with integer legs $a,b$ and hypotenuse $c=b+1$, where $b<100$. The Pythagorean theorem states $a^2+b^2=c^2$. Substituting $c=b+1$: \newline
  $a^2+b^2=(b+1)^2$ \newline
  $a^2+b^2=b^2+2b+1$ \newline
  $a^2=2b+1$ \newline
  This implies $2b+1$ must be a perfect square. Let $a^2=k^2$ for some integer $a$. So $2b+1=a^2 \implies 2b=a^2-1 \implies b=\frac{a^2-1}{2}$. Since $b$ must be an integer, $a^2-1$ must be even, which means $a^2$ must be odd, so $a$ must be odd. Also, we are given $b<100$. Substituting $b$: \newline
  $\frac{a^2-1}{2}<100$ \newline
  $a^2-1<200$ \newline
  $a^2<201$ \newline
  Since $a$ is a positive integer (leg length), $a\geq1$. Possible values for $a$: $1,3,5,\dots$ such that $a^2<201$. $\sqrt{201}\approx14.17$. So $a$ can be $1,3,5,7,9,11,13$. Let's verify for each $a$: \newline
  1. $a=1 \implies b=(1-1)/2=0$. Usually leg lengths are positive integers. If $b=0$, it's a degenerate triangle. The problem asks for ``right triangles,'' implying non-degenerate. So $b>0 \implies a^2>1 \implies a>1$. \newline
  2. $a=3 \implies b=(9-1)/2=4$. $c=5$. $3^2+4^2=5^2$. Valid. \newline
  3. $a=5 \implies b=(25-1)/2=12$. $c=13$. Valid. \newline
  4. $a=7 \implies b=(49-1)/2=24$. $c=25$. Valid. \newline
  5. $a=9 \implies b=(81-1)/2=40$. $c=41$. Valid. \newline
  6. $a=11 \implies b=(121-1)/2=60$. $c=61$. Valid. \newline
  7. $a=13 \implies b=(169-1)/2=84$. $c=85$. Valid. \newline
  Next odd $a=15 \implies a^2=225>201$, so $b>100$. \newline
  So the valid values for $a$ are $3,5,7,9,11,13$. This gives 6 solutions. I will write a python script to verify this count. \newline
  \texttt{</tool\_response> assistant} \\
\addlinespace[3pt]
\multicolumn{3}{r@{}}{\textit{Panel (a) continues on the next page.}} \\
\bottomrule
\end{tabular}
\end{table*}

\begin{table*}[p]
\centering
\scriptsize
\setlength{\tabcolsep}{3.5pt}
\renewcommand{\arraystretch}{1.08}
\begin{tabular}{@{}>{\raggedright\arraybackslash}p{0.07\textwidth}>{\raggedright\arraybackslash}p{0.14\textwidth}>{\raggedright\arraybackslash}p{0.74\textwidth}@{}}
\toprule
\multicolumn{3}{@{}l}{\textbf{Table~\ref{tab:stage2-conversion-example} (continued): (a) Original Stage-I collaborative trajectory (Part 2 of 2)}} \\
\addlinespace[2pt]
\textbf{Turn} & \textbf{Producer} & \textbf{Original serialized content} \\
\midrule

Call 2
& Controller
& \cellcolor{orange!8}\texttt{<tool\_call>} \newline
  \texttt{\{"name": "code\_interpreter", "arguments": \{"model": "Qwen3.5-9B",} \newline
  \texttt{"instruction": "Verify the number of valid right triangles with integer legs a and b, hypotenuse b+1, and b < 100. Compute the valid odd values of a such that a\^{}2 < 201 and b = (a\^{}2 - 1)/2 > 0 and b < 100.",} \newline
  \texttt{"code": ""\}\}} \newline
  \texttt{</tool\_call>} \\
\addlinespace[2pt]

Code
& Qwen3.5-9B
& \cellcolor{orange!8}\texttt{user <tool\_response> <python>} \newline
  \texttt{count = 0} \newline
  \texttt{valid\_triangles = []} \newline
  \texttt{for a in range(1, 100):} \newline
  \quad\texttt{b = (a**2 - 1) / 2} \newline
  \quad\texttt{if b.is\_integer() and b > 0 and int(b) < 100:} \newline
  \qquad\texttt{count += 1} \newline
  \qquad\texttt{valid\_triangles.append((a, int(b), int(b)+1))} \newline
  \texttt{print(f"Number of valid triangles: \{count\}")} \newline
  \texttt{for t in valid\_triangles:} \newline
  \quad\texttt{print(t)} \newline
  \texttt{</python>} \\

Output
& Python sandbox
& \cellcolor{black!5}\texttt{<output>} \newline
  \texttt{Number of valid triangles: 6} \newline
  \texttt{(3, 4, 5)} \newline
  \texttt{(5, 12, 13)} \newline
  \texttt{(7, 24, 25)} \newline
  \texttt{(9, 40, 41)} \newline
  \texttt{(11, 60, 61)} \newline
  \texttt{(13, 84, 85)} \newline
  \texttt{</output>} \\
\addlinespace[2pt]

Final
& Controller
& \cellcolor{green!6}\texttt{</tool\_response> assistant} \newline
  The verification confirms there are 6 valid right triangles with integer leg lengths $a$ and $b$, hypotenuse $b+1$, and $b<100$. The triangles are: \newline
  -- $(3,4,5)$ \newline
  -- $(5,12,13)$ \newline
  -- $(7,24,25)$ \newline
  -- $(9,40,41)$ \newline
  -- $(11,60,61)$ \newline
  -- $(13,84,85)$ \newline
  All conditions are satisfied, and no degenerate or invalid cases are included. \newline
  \texttt{Answer: \textbackslash boxed\{6\}} \\
\bottomrule
\end{tabular}
\end{table*}

\begin{table*}[p]
\centering
\scriptsize
\setlength{\tabcolsep}{3.5pt}
\renewcommand{\arraystretch}{1.00}
\begin{tabular}{@{}>{\raggedright\arraybackslash}p{0.065\textwidth}>{\raggedright\arraybackslash}p{0.13\textwidth}>{\raggedright\arraybackslash}p{0.685\textwidth}c@{}}
\toprule
\multicolumn{4}{@{}l}{\textbf{Table~\ref{tab:stage2-conversion-example} (continued): (b) Complete converted Stage-II training sequence}} \\
\addlinespace[2pt]
\textbf{Turn} & \textbf{Stage-I source} & \textbf{Serialized content after canonicalization} & \makecell{\textbf{Target}\\\textbf{mask}} \\
\midrule

Input
& Problem prompt
& Solve the following math problem step by step. The last line of your response should be exactly of the form \texttt{Answer: \textbackslash boxed\{answer\}}, where \texttt{answer} is the answer to the problem. \newline
  Determine how many right triangles exist with integer leg lengths $a$ and $b$, where the hypotenuse is of length $b+1$ and $b<100$. \newline
  Remember to put your final answer on its own line using exactly: \texttt{Answer: \textbackslash boxed\{answer\}}.
& \textbf{0} \\
\addlinespace[2pt]

Call 1
& Controller
& \cellcolor{blue!3}\texttt{<tool\_call>} \newline
  \texttt{\{"name": "think", "arguments": \{"model": "Qwen3.5-9B",} \newline
  \texttt{"instruction": "Solve the current problem independently and provide rigorous mathematical reasoning."\}\}} \newline
  \texttt{</tool\_call>}
& \textbf{1} \\
\addlinespace[2pt]

Response 1
& Qwen3.5-9B
& \cellcolor{blue!3}\texttt{user <tool\_response>} \newline
  I need to find the number of right triangles with integer legs $a,b$ and hypotenuse $c=b+1$, where $b<100$. The Pythagorean theorem states $a^2+b^2=c^2$. Substituting $c=b+1$: \newline
  $a^2+b^2=(b+1)^2$ \newline
  $a^2+b^2=b^2+2b+1$ \newline
  $a^2=2b+1$ \newline
  This implies $2b+1$ must be a perfect square. Let $a^2=k^2$ for some integer $a$. So $2b+1=a^2 \implies 2b=a^2-1 \implies b=\frac{a^2-1}{2}$. Since $b$ must be an integer, $a^2-1$ must be even, which means $a^2$ must be odd, so $a$ must be odd. Also, we are given $b<100$. Substituting $b$: \newline
  $\frac{a^2-1}{2}<100$ \newline
  $a^2-1<200$ \newline
  $a^2<201$ \newline
  Since $a$ is a positive integer (leg length), $a\geq1$. Possible values for $a$: $1,3,5,\dots$ such that $a^2<201$. $\sqrt{201}\approx14.17$. So $a$ can be $1,3,5,7,9,11,13$. Let's verify for each $a$: \newline
  1. $a=1 \implies b=(1-1)/2=0$. Usually leg lengths are positive integers. If $b=0$, it's a degenerate triangle. The problem asks for ``right triangles,'' implying non-degenerate. So $b>0 \implies a^2>1 \implies a>1$. \newline
  2. $a=3 \implies b=(9-1)/2=4$. $c=5$. $3^2+4^2=5^2$. Valid. \newline
  3. $a=5 \implies b=(25-1)/2=12$. $c=13$. Valid. \newline
  4. $a=7 \implies b=(49-1)/2=24$. $c=25$. Valid. \newline
  5. $a=9 \implies b=(81-1)/2=40$. $c=41$. Valid. \newline
  6. $a=11 \implies b=(121-1)/2=60$. $c=61$. Valid. \newline
  7. $a=13 \implies b=(169-1)/2=84$. $c=85$. Valid. \newline
  Next odd $a=15 \implies a^2=225>201$, so $b>100$. \newline
  So the valid values for $a$ are $3,5,7,9,11,13$. This gives 6 solutions. I will write a python script to verify this count. \newline
  \texttt{</tool\_response> assistant}
& \textbf{1} \\
\addlinespace[2pt]

Call 2
& Controller
& \cellcolor{orange!8}\texttt{<tool\_call>} \newline
  \texttt{\{"name": "code\_interpreter", "arguments": \{"model": "Qwen3.5-9B",} \newline
  \texttt{"instruction": "Solve or verify the current problem using Python when helpful.", "code": ""\}\}} \newline
  \texttt{</tool\_call>}
& \textbf{1} \\
\addlinespace[2pt]

Code
& Qwen3.5-9B
& \cellcolor{orange!8}\texttt{user <tool\_response> <python>} \newline
  \texttt{count = 0} \newline
  \texttt{valid\_triangles = []} \newline
  \texttt{for a in range(1, 100):} \newline
  \quad\texttt{b = (a**2 - 1) / 2} \newline
  \quad\texttt{if b.is\_integer() and b > 0 and int(b) < 100:} \newline
  \qquad\texttt{count += 1} \newline
  \qquad\texttt{valid\_triangles.append((a, int(b), int(b)+1))} \newline
  \texttt{print(f"Number of valid triangles: \{count\}")} \newline
  \texttt{for t in valid\_triangles:} \newline
  \quad\texttt{print(t)} \newline
  \texttt{</python>}
& \textbf{1} \\

Output
& Python sandbox
& \cellcolor{black!5}\texttt{<output>} \newline
  \texttt{Number of valid triangles: 6} \newline
  \texttt{(3, 4, 5)} \newline
  \texttt{(5, 12, 13)} \newline
  \texttt{(7, 24, 25)} \newline
  \texttt{(9, 40, 41)} \newline
  \texttt{(11, 60, 61)} \newline
  \texttt{(13, 84, 85)} \newline
  \texttt{</output>}
& \textbf{0} \\
\addlinespace[2pt]

Final
& Controller
& \cellcolor{green!6}\texttt{</tool\_response> assistant} \newline
  The verification confirms there are 6 valid right triangles with integer leg lengths $a$ and $b$, hypotenuse $b+1$, and $b<100$. The triangles are: \newline
  -- $(3,4,5)$ \newline
  -- $(5,12,13)$ \newline
  -- $(7,24,25)$ \newline
  -- $(9,40,41)$ \newline
  -- $(11,60,61)$ \newline
  -- $(13,84,85)$ \newline
  All conditions are satisfied, and no degenerate or invalid cases are included. \newline
  \texttt{Answer: \textbackslash boxed\{6\}}
& \textbf{1} \\
\bottomrule
\end{tabular}

\end{table*}

\end{document}